\documentclass[letterpaper]{article} 
\usepackage{aaai25}  
\usepackage{times}  
\usepackage{helvet}  
\usepackage{courier}  
\usepackage[hyphens]{url}  
\usepackage{graphicx} 
\usepackage{natbib}  
\usepackage{caption} 
\usepackage{algorithm}
\usepackage{algorithmic}

\usepackage{newfloat}
\usepackage{listings}
\DeclareCaptionStyle{ruled}{labelfont=normalfont,labelsep=colon,strut=off} 
\floatstyle{ruled}
\newfloat{listing}{tb}{lst}{}
\floatname{listing}{Listing}
\usepackage{amsmath}
\usepackage{amssymb}
\usepackage{booktabs}
\usepackage{subcaption}

\title{Evaluate with the Inverse: \\ Efficient Approximation of Latent Explanation Quality Distribution
}
\author{
    Carlos Eiras-Franco\textsuperscript{\rm 1}\\
    Anna Hedström\textsuperscript{\rm 2,}\textsuperscript{\rm 3,}\textsuperscript{\rm 5},
    Marina M.-C. Höhne\textsuperscript{\rm 4,}\textsuperscript{\rm 5}
}
\affiliations{
    \textsuperscript{\rm 1}Universidade da Coruña. CITIC\\

    \textsuperscript{\rm 2}UMI Lab, Leibniz Institute of Agricultural Engineering and Bioeconomy e.V. (ATB)\\

    \textsuperscript{\rm 3}BIFOLD – Berlin Institute for the Foundations of Learning and Data\\

    \textsuperscript{\rm 4}Data Science Department, Leibniz Institute of Agricultural Engineering and Bioeconomy e.V. (ATB)\\

    \textsuperscript{\rm 5}Department of Computer Science, University of Potsdam\\

    carlos.eiras.franco@udc.es
}

\begin{document}

\maketitle

\begin{abstract}
Obtaining high-quality explanations of a model's output enables developers to identify and correct biases, align the system's behavior with human values, and ensure ethical compliance. Explainable Artificial Intelligence (XAI) practitioners rely on specific measures to gauge the quality of such explanations. These measures assess key attributes, such as how closely an explanation aligns with a model's decision process (faithfulness), how accurately it pinpoints the relevant input features (localization), and its consistency across different cases (robustness). Despite providing valuable information, these measures do not fully address a critical practitioner's concern: how does the quality of a given explanation compare to other potential explanations?
Traditionally, the quality of an explanation has been assessed by comparing it to a randomly generated counterpart. This paper introduces an alternative: the Quality Gap Estimate (\texttt{QGE}). The \texttt{QGE} method offers a direct comparison to what can be viewed as the `inverse' explanation, one that conceptually represents the antithesis of the original explanation.
Our extensive testing across multiple model architectures, datasets, and established quality metrics demonstrates that the \texttt{QGE} method is superior to the traditional approach. Furthermore, we show that \texttt{QGE} enhances the statistical reliability of these quality assessments. This advance represents a significant step toward a more insightful evaluation of explanations that enables a more effective inspection of a model's behavior.
\end{abstract}

%

\section{Introduction}
\label{sec:Introduction}

\begin{figure*}[!t]
    \centering
    \includegraphics[width=\linewidth]{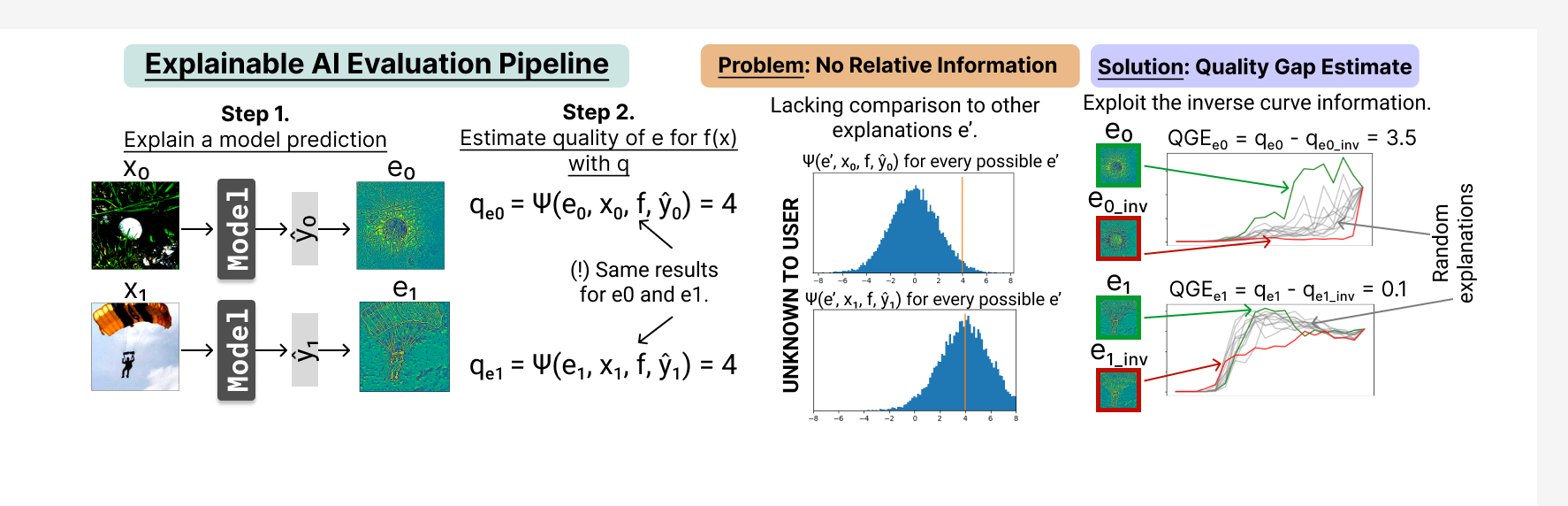}
    \caption{(Step 1). The usual XAI pipeline allows the user to obtain an explanation $e$ for a prediction $\hat{y}$ using any explanation method. This is demonstrated for two distinct inputs ($\mathbf{x_0}$ and $\mathbf{x_1}$), producing predictions $\hat{y}_0$ and $\hat{y}1$, and explanations $\mathbf{e_0}$ and $\mathbf{e_1}$, respectively. (Step 2). To assess the quality of explanation $\mathbf{e}$ for prediction $\hat{y}$, the user computes a quality measure $q$. In this example, we use the area under the Pixel-Flipping curve, though the method can work with any attribution-based quality measure. (Problem) Despite both input/explanation pairs registering identical $q$ values, it remains unknown to the user that $\mathbf{e_0}$ has higher quality than most explanations for the first prediction, while $\mathbf{e_1}$ has average quality compared to other explanations for the second prediction, as shown by their histograms. (Solution) To allow the user to effectively gauge the relative quality of explanation $\mathbf{e}$ against alternative explanations $\mathbf{e'}$, we introduce $\texttt{QGE}$, which measures the difference between the quality of $\mathbf{e}$ and the quality of $\mathbf{e}_{inv}$ (a rearrangement of $\mathbf{e}$ ranking features in reverse order). This comparative quality metric does not require costly sampling of the $q$ distribution. Although both explanations have equivalent $q$ values, using \texttt{QGE}, the user can discern that $\mathbf{e_0}$ is a high-quality explanation for the first prediction, while $\mathbf{e_1}$ is merely average for the second one. The user may then seek a better explanation for the second prediction.}
    \label{fig:general_explanation}
\end{figure*}

Model output explanations play a crucial role in AI alignment by enhancing transparency, understandability, and trustworthiness in AI systems. In most contexts related to explainability, ground-truth explanations are unavailable \cite{dasgupta2022, hedstrom2023metaquantus}. This absence inherently complicates the task of evaluating explanations. Consequently, efforts to evaluate explanations vary widely, ranging from assessing the robustness of explanations to noise, their complexity, and their localization, to evaluating how faithfully an explanation represents the underlying model.

Although it is not possible to develop a metric based on verified ground truth, a crucial insight is that the adequacy of an explanation can still be assessed by comparing it relative to other explanations with the use of quality measures. Commonly, XAI evaluation methods yield numerical fitness measures that indicate the degree to which an explanation adheres to a certain criterion. However, these numerical values do not provide insight into an explanation's standing relative to the spectrum of possible explanations. In this study, we introduce a method that quantifies the quality of attribution-based explanation methods by efficiently estimating how an explanation's quality compares to the rest of the potential explanations. Despite its apparent simplicity, our evaluations using various sanity checks demonstrate that this strategy enhances the reliability of quality metrics.

While other studies have employed different notions of randomness to compare or evaluate explanation methods \cite{findingrightbommer, samek2015, adebayo2018}, these approaches are often more computationally intensive and yield less favorable results. The proposed method is independent of the specific dataset, model, task, and, importantly, of the quality metric used.

The main contributions of this paper include:
\begin{itemize}
    \item The introduction of the Quality Gap Estimate (\texttt{QGE}), a novel evaluation strategy that renders existing quality metrics more informative by aiding in the determination of an explanation’s quality relative to alternatives with a limited increase of computational requirements.
    \item A redefinition of the problem of assessing the superiority of an explanation over its alternatives as a sampling problem, which generalizes the conventional method of comparison with a random explanation.
    \item An assessment of the applicability of \texttt{QGE} across a wide array of established quality metrics, evaluating its impact on various dimensions including faithfulness, localization, and robustness.
    \item The presentation of experimental findings that demonstrate an enhancement in the statistical reliability of existing quality metrics through the application of \texttt{QGE}, providing XAI practitioners with more reliable interpretation tools.
\end{itemize}
\section{Related Works}

Evaluating the quality of methods without ground truth explanation labels is a significant challenge for reserachers~\cite{brunke, brocki2022, rong2022, hedstrom2023metaquantus}. Many studies have shown that faithfulness metrics are highly sensitive to their parameterisation during evaluation; altering patch sizes or pixel occlusion tactics can significantly affect the outcomes~\cite{tomsett2020, brunke, brocki2022, rong2022, hedstrom2023metaquantus}. 
These findings are concerning; if small changes in parameters cause large variations in evaluation outcomes, it may be hard to trust the results. 
 
For more reliable estimations of explanation quality, individual explanation methods have been evaluated \textit{relative} to a random baseline~\cite{samek2015, nguyen2020, ancona2019}. The concept of using the random explainer as a worst-case reference point has also been used for calculating explanation skill scores~\citep{findingrightbommer} or as part of a paired t-test to compare with existing explanation methods~\citep{irof2020}. Most similar to our work~\cite{blücher2024decoupling} is incorporating information about the pixel-flipping inverse curve. Our contribution is different in both aim and applicability. Their approach aims at enhancing the occlusion process, specifically the masking strategy for pixel-flipping~\cite{samek2015}. We provide a general-purpose evaluative solution applicable across various explanation metrics such as localization, faithfulness, and robustness.

\section{Method}
\label{sec:Method}

\subsection{A Framework for Explanation Quality}

Consider a classification problem where we have a model, denoted as a function $f$ that maps an input $\mathbf{x}\in\mathbb{R}^D$ to an output $\hat{y}\in \{1, \ldots, C\}$. This function estimates the probability distribution across classes i.e., $f(\mathbf{x})_i=p(y_i|\mathbf{x})~\forall i \in \{1, \ldots, C\}$, so $\hat{y} = \text{argmax}(f(\mathbf{x}))$. To identify the features utilized by the model to predict $\hat{y}$, we employ an explanation function $\Phi$ as follows:
\begin{equation}
    \Phi(\mathbf{x}, f, \hat{y}) = \mathbf{e}
\end{equation}
$\Phi$ outputs a real-valued vector $\mathbf{e}$ with $D$ components that assign attribution to each feature $x_i$ in $\mathbf{x}$, indicating its relative importance in $f$'s prediction of class $\hat{y}$.

Various methods exist to assess the suitability or fitness of $\mathbf{e}$ based on different attributes. Generally, we can define a quality measure $\Psi$ as a function that evaluates the quality of a given explanation $\mathbf{e}$, relative to the model $f$, the input $\mathbf{x}$, and the predicted label $\hat{y}$. For brevity, we denote this scalar value as $q_e$, and we use simply $q$ to refer to the function with a fixed $f$, $\mathbf{x}$, and $\hat{y}$ that evaluates a given explanation:

\begin{equation}
    q_e := q(\mathbf{e}) := \Psi(\mathbf{e}, \mathbf{x}, f, \hat{y})
    \label{eq:q_e}
\end{equation}


\subsection{The Need for Relative Information}
\label{sec:Method_relative}
The value $q_e$ provides practitioners with a measure of how well an explanation $\mathbf{e}$ adheres to predefined quality criteria. It enables comparisons between different explanations by contrasting $q_e$ with $q_{e'}$. However, a crucial question remains for practitioners: How good is $\mathbf{e}$ compared to the whole set of alternative explanations? (see Figure~\ref{fig:general_explanation}).

To answer this question, we could check the position of $q_e$ in the unknown distribution of $q$ across all possible explanations $\mathbf{e'}$. Yet, this approach is impractical as it requires calculating the distribution of $q$ for all possible explanations, which is computationally infeasible.

A strategy to tackle this problem consists of estimating the latent distribution of the quality measure by sampling $q$ values. For instance, Pixel-Flipping \cite{bach2015pixel} compares $q_{e}$ against $q_{e^r}$ which represents the quality score for a random explanation (denoted $\mathbf{e}^r$). This method effectively performs a single-sample estimation of the quality distribution. Pixel-Flipping transforms the original quality measure $q$ into a new transformed measure:

\begin{equation}
    qt(\mathbf{e}) = \Psi(\mathbf{e}, \mathbf{x}, f, \hat{y}) - \Psi(\mathbf{e^r}, \mathbf{x}, f, \hat{y}) = q_e - q_{e^r}
    \label{eq:qt_pixel_flipping}
\end{equation}

However, relying on just one random sample ($q_{e^r}$) can compromise the accuracy of the estimation. A more robust transformation would involve sampling multiple random explanations, computing their quality, and calculating an average. Yet, this method incurs higher computational costs.

To overcome these challenges, our goal is to develop a transformed quality measure $qt$ that satisfies the following criteria:
\begin{itemize}
    \setlength\itemindent{1em} 
    \item[\textbf{(R1)}] Provides a value that clearly indicates the relative standing of $q_{e}$ within the latent distribution of all possible $q_{e'}$ values. By evaluating $qt(\mathbf{e})$, a user should be able to determine whether $\mathbf{e}$'s quality is above-average, average, or below-average.
    \item[\textbf{(R2)}] Preserves the comparative information inherent in $q$, especially its capacity to rank explanations. Given a explanation $\mathbf{e^i}$, any explanation $\mathbf{e^j}$ with higher quality, should also have a higher $qt$. More formally, $qt$ should be constructed so that, given any pair of explanations $(e^i, e^j)$, if $q_{e^i}<q_{e^j}$ holds, then $qt(\mathbf{e}^i)<qt(\mathbf{e}^j)$.
    \item[\textbf{(R3)}] Is computationally efficient.
\end{itemize}

\subsection{Proposed Method}\label{sec:Proposed_method}
We propose using the $\mathbf{e}^{inv}$ explanation, which ranks features in inverse order to $\mathbf{e}$, as an alternative to the commonly used random explanation $\mathbf{e}^r$. This approach aims to improve the quality of estimations while maintaining low computational cost. Given $o = \text{argsort}(\mathbf{e})$, we define
\begin{equation}
\label{eq:e_inv}
    e^{inv}_{o_i} := e_{o_{D-i+1}}~~\forall{i} \in \left[1..D\right]
\end{equation}
where $D$ is the number of variables in $\mathbf{e}$.
Therefore, $\mathbf{e}^{inv}$ is a permutation of the values of $\mathbf{e}$ constructed so that the most attributed variable in $\mathbf{e}$ gets the smallest attribution in $\mathbf{e}^{inv}$, the second most attributed variable in $\mathbf{e}$ gets the second smallest attribution, and so on. $\mathbf{e}^{inv}$ is, then, the opposite interpretation of the original explanation $\mathbf{e}$. An example of this procedure is shown below.

  \begin{minipage}{\columnwidth}
    \begin{align*}
    \mathbf{e} &= [0.1, -0.1, 9.0, 4.0]  & o=\text{argsort}(\mathbf{e}) &= [1, 0, 3, 2] \\
    \mathbf{e}^{inv} &= [4.0, 9.0, -0.1, 0.1]  & \text{argsort}(\mathbf{e}^{inv}) &= [2, 3, 0, 1]
    \end{align*}
  \end{minipage}
By design, the ranking by attribution of the features in $\mathbf{e}^{inv}$ is the same as for $\mathbf{e}$, but in reverse order (i.e. $\text{argsort}(\mathbf{e})$ = $\text{reversed}(\text{argsort}(\mathbf{e}^{inv}))$).

Once we have $\mathbf{e^{inv}}$, we define the proposed transformation, which we will call its Quality Gap Estimation (\texttt{QGE}), as the difference between the quality value of the original explanation $\mathbf{e}$ and the quality value of $\mathbf{e^{inv}}$:

\begin{equation}
\label{eq:qge}
    \texttt{QGE} = \Psi(\mathbf{e}, \mathbf{x}, f, \hat{y}) - \Psi(\mathbf{e^{inv}}, \mathbf{x}, f, \hat{y})
\end{equation}

The rationale behind this method is intuitive: \texttt{QGE} increases not only when $\Psi(\mathbf{e}, \mathbf{x}, f, \hat{y})$ is high, indicating the high quality of $\mathbf{e}$, but also when $\Psi(\mathbf{e}^{inv}, \mathbf{x}, f, \hat{y})$ is low, reflecting the poor quality of the inversely ranked explanation. A substantial gap between these values suggests that many random explanations would have quality values falling between these two, thereby indicating that $\mathbf{e}$ is of much higher quality than an average-quality explanation. Conversely, a small \texttt{QGE} implies that the quality difference between $\mathbf{e}$ and $\mathbf{e}^{inv}$ is minimal, suggesting that the order in which $\mathbf{e}$ ranks features is approximately as effective as any alternative, regardless of the absolute value of $q_e$. This pattern suggests that \texttt{QGE} satisfies requirement \textbf{R1}, with values approximately zero for average-quality explanations, negative for below-average, and positive for above-average explanations. Moreover, if \textbf{R2} is met, the more above-average $\mathbf{e}$'s quality is, the higher \texttt{QGE} will be, and similarly, the more below-average $q_e$'s quality is, the lower \texttt{QGE} will be. The level of compliance with \textbf{R2} is assessed in Section \ref{sec:suitability_qge}.

Regarding \textbf{R3}, this requirement is met because \texttt{QGE} can be computed quickly. Generally, the cost of computing any transformation $qt$ is dominated by the cost of computing $\Psi$, which is generally a costly function. The fewer times a transformation $qt$ needs to compute $\Psi$, the faster it will be. Determining the \texttt{QGE} requires computing $\Psi$ only twice (the original $\Psi(\mathbf{e}, \mathbf{x}, f, \hat{y})$, and the additional $\Psi(\mathbf{e^{inv}}, \mathbf{x}, f, \hat{y})$). This approach avoids the computational expense of needing multiple samples of $\Psi$ for different explanations to estimate a distribution, as alternative methods do. Other than computing $\Psi$, \texttt{QGE} requires a single subtraction and computing $\mathbf{e}^{inv}$ as indicated in Eq.~\ref{eq:e_inv}. While the time needed to compute $\mathbf{e}^{inv}$ is generally negligible compared to the cost of computing $\Psi$, this step can also be sped up in most cases. Although Eq.~\ref{eq:e_inv} preserves the original attribution values and reallocates them in reverse order, an alternate but straightforward computation could set $e_i^{inv} := -e_i \forall i \in [0..D]$ to achieve the goal of inversely ranking features compared to $\mathbf{e}$ ($\text{argsort}(e)$ = $\text{reversed}(\text{argsort}(e^{inv}))$). However, the magnitudes of the attributions are not maintained after this transformation and, in instances where the quality metric $\Psi$ requires bounded attribution values, users would need to adhere to the original formulation in Equation \ref{eq:e_inv} if a shift and scale of $e_i^{inv} := -e_i$ is unsuitable.

\ifdefined\doubleblind
  An implementation of \texttt{QGE} is available in [GitHub repository removed for double-blind review]
\else
  An implementation of \texttt{QGE} for a wide variety of quality measures is available in the Quantus toolkit: \url{https://github.com/understandable-machine-intelligence-lab/Quantus}
\fi

\section{Experimental Results}\label{sec:experimental-results}
To verify the quality of our method we performed experiments focused on two main points:

\begin{itemize}
    \item Assesing the level of compliance of \texttt{QGE} with the requirements listed in Section~\ref{sec:Method_relative}:
    \begin{enumerate}
        \item[\textbf{(R1)}] is met by the design of \texttt{QGE} (see intuitive explanation after Eq.~{\ref{eq:qge}}). We report a complete exploration of the \texttt{QGE} distribution that confirms this.
        \item[\textbf{(R2)}] Evaluating \texttt{QGE} in its ability to preserve the information inherent in the original quality metric $q$ (\textbf{R2}). The details of this evaluation are discussed in Section~\ref{sec:suitability_qge}.
        \item[\textbf{(R3)}] The cost of computing a transformation $qt$ is dominated by how many times it needs to compute the original quality function $\Psi$. Our experiments confirm this, and the results presented allow the comparison of \texttt{QGE} with an alternative of similar cost. 
    \end{enumerate}

    \item Assessing the statistical reliability of \texttt{QGE} compared to competitive baseline methods. The experiments conducted for this are reported in Section \ref{sec:effect_on_metrics}. 
\end{itemize}

In our experiments, we took a fixed model $f$, input sample $\mathbf{x}$, and label $y$. Although these parameters remained constant for each individual experiment, we varied them across different experiments to test the robustness and general applicability of our method.

\ifdefined\doubleblind
  The code used for all experiments is available at [GitHub repository removed for double-blind review]
\else
  The code used for all experiments is available at \url{https://github.com/annahedstroem/eval-project}
\fi

\subsection{Suitability of \texttt{QGE}}\label{sec:suitability_qge}
For our initial experiments, we employed the \texttt{Pixel-Flipping} quality measure \cite{bach2015pixel}. Let $f_y(\mathbf{x})$ represent the output of model $f$ for class $y$, and $\mathbb{P}(\mathbf{x},\mathbf{e}, M)$ denote a perturbation function that modifies all features of $\mathbf{x}$ except for the $M$ most attributed features according to explanation $\mathbf{e}$ (with $M \in [0,D]$). The quality of explanation $\mathbf{e}$ is then measured as the average value\footnote{Some implementations measure the area under the curve instead of the average activation level. Both alternatives are equivalent since they are proportional.} of $f_y(\mathbf{x})$ over all possible levels of feature selection $M$:

\begin{equation}
\label{eq:q_value}
    q_e:=\Psi(\mathbf{e},\mathbf{x},f,y) = \frac{1}{D}\sum_{m=0}^{D}{f_y(\mathbb{P}(\mathbf{x},\mathbf{e},m))}
\end{equation}

In this experimental setup, the perturbation function $\mathbb{P}(\mathbf{x}, \mathbf{e}, M)$ results in $\mathbf{x'}$, where the $D-M$ least relevant features (as determined by $\mathbf{e}$) are replaced with zeros. This introduces the well-known problem that $\mathbf{x'}$ is outside the distribution for which the model $f$ was trained~\cite{hase2021}, an issue that is addressed below (see ``Advantage of using \texttt{QGE} vs. $\texttt{QRAND}_1$ across datasets and models").

We compared two different transformations: our evaluation measure, \texttt{QGE}, was compared against $\texttt{QRAND}_{\text{K}}$, a baseline measure that estimates the quality distribution across all explanations by sampling $k$ random explanations. It calculates the difference between $q_e$ and the average quality of those $k$ samples, generalizing the usual comparison with a single random explanation.
\begin{equation}
    \label{eq:qrand}
    \texttt{QRAND}_{\text{K}}
    = q_{e} - \frac{\sum_{i=0}^{K}{q_{r_k}}}{K}
\end{equation} where ${q_{r_k}}$ represents the quality of a random explanation.

To assess adherence to \textbf{R2} for both quality metric transformations $qt$ (either \texttt{QGE} or $\texttt{QRAND}_{\text{K}}$ for a range of $k$ values), we used the following evaluation metrics (both implemented in SciPy~\cite{2020SciPy-NMeth}):
\begin{itemize}
    \item \textbf{Kendall rank correlation ($\tau$):} As a direct translation of \textbf{R2}, we computed Kendall's rank correlation $\tau$, i.e. the level of agreement of the order after the transformation with the order after the transformation (i.e. $q_i<q_j \Longrightarrow qt_i<qt_j$ and $q_i>q_j \Longrightarrow qt_i>qt_j$).
    \item \textbf{Spearman correlation ($\rho$)} between the transformed quality metric $qt$ (either \texttt{QGE} or $\texttt{QRAND}_{\text{K}}$) and the original $q$.
\end{itemize}

Four scenarios are reported in Section \ref{sec:exhaustive_exploration}: (a) the examination of the complete set of all possible explanations for two small datasets; (b) an exploration in larger datasets; (c) an analysis of the effect of the model used; and (d) the use of measures other than \texttt{Pixel-Flipping}.

\subsubsection{a. Exhaustive exploration of the explanation space}\label{sec:exhaustive_exploration}
We first tested our method using two small datasets in which the complete list of all possible explanations can be enumerated in a reasonable time. It's crucial to note that differences in $q_e$ and $q_{e'}$ arise solely when $\mathbf{e}$ and $\mathbf{e'}$ differ in their feature ranking orders. This constraint significantly simplifies the space of possible explanations and facilitates the enumeration process.

The datasets selected for this experiment were the \texttt{Avila} dataset~\cite{avila_dataset} and the \texttt{Glass Identification} dataset~\cite{misc_glass_identification_42}, both small tabular datasets. Given the limited number of variables in these datasets, the complete set of distinct explanations (with respect to $q$; i.e. every attribution vector that yields a different order when argsorted) corresponds to the set of all possible permutations of the variables. This set can be exhaustively explored. The examples in \texttt{Avila} consist of 10 variables, which yields a total of $10!=3,628,800$ different explanations while \texttt{Glass} has 9 variables, amounting to $9!=362,880$ explanations.

Our analysis confirms that the distribution of \texttt{QGE} is centered on zero (see Appendix~\ref{sec:distribution-qge}). Consequently, if \textbf{R2} is met, (i.e. high-quality explanations obtain higher \texttt{QGE} values), then \textbf{R1} is also met since the user can clearly distinguish between explanations with an above-average quality (which have positive \texttt{QGE}) and explanations with below-average quality (which have negative \texttt{QGE}). The following experiments aim to assess the degree to which \textbf{R2} is met.

For each dataset, we trained a Multilayer Perceptron (\texttt{MLP}). The models achieved test accuracies of 0.99 on the \texttt{Avila} dataset and 0.77 on the \texttt{Glass} dataset, respectively. For each explanation $\mathbf{e}$ in the set of all possible explanations, we calculated $q_e$ (Eq.\ref{eq:q_value}), \texttt{QGE} (Eq. \ref{eq:e_inv}), and $\texttt{QRAND}_{\text{K}}$ (Eq. \ref{eq:qrand}) for values of $k$ ranging from 1 to 10.


\paragraph{Order preservation: Kendall's $\tau$}

To measure the extent to which a transformation $qt$ (either $\texttt{QRAND}_{\text{K}}$ or \texttt{QGE}) preserves the order of the original quality measure $q$, we compute Kendall's $\tau_{q,qt}$. 

\begin{figure}[htbp]
    \centering
    \includegraphics[width=0.48\linewidth]{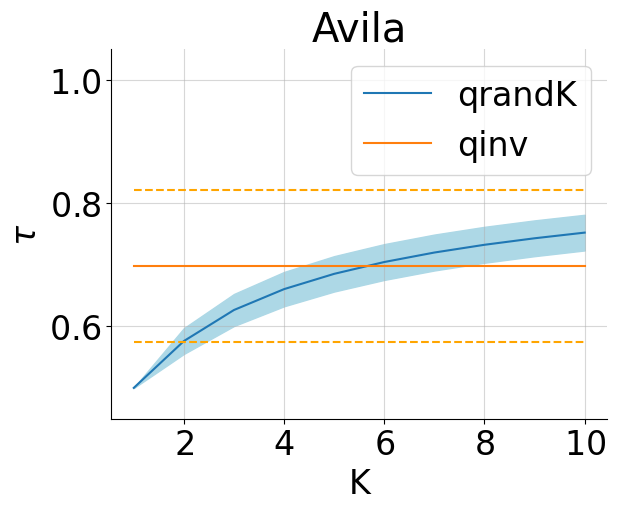}
    \includegraphics[width=0.48\linewidth]{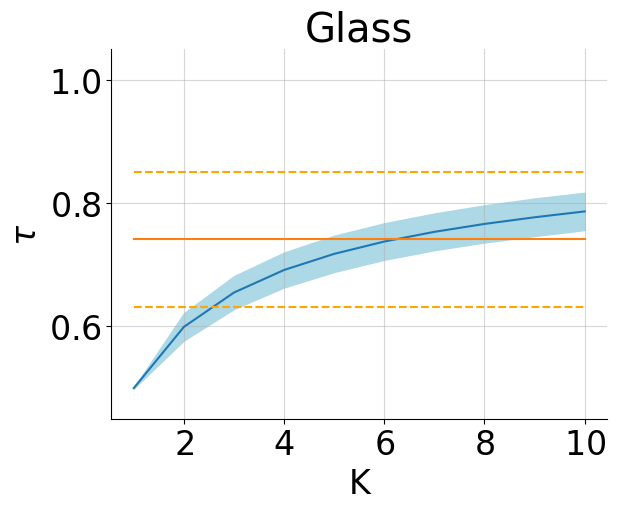}
    \caption{Kendall's $\tau$ for the \texttt{Avila} and \texttt{Glass} datasets. The blue line indicates the average $\tau_{q,\texttt{QRAND}_{\text{K}}}$ for each value of $K$ over 5 different inputs, with the shaded area showing the average 
 $\pm\sigma$. The orange line records the average $\tau_{q,\texttt{QGE}}$, with dashed lines representing the average $\pm\sigma$.)}
    \label{fig:exhaustive_correct_pairs}
\end{figure}

In Fig.~\ref{fig:exhaustive_correct_pairs} we report the results for 5 different $\mathbf{x}$ inputs, for each of which all possible explanations were computed. These show that for \texttt{Avila}, on average, using \texttt{QGE} as a transformation maintains the correct ordering for 85\% of pairs, which results in a $\tau_{q,\texttt{QGE}}$ of 0.7 ($\pm$ 0.12). To obtain a comparable result with the conventional $\texttt{QRAND}_{\text{K}}$, more than $k=6$ random samples are needed. Similarly, for the \texttt{Glass} dataset, \texttt{QGE} obtains a $\tau_{q,\texttt{QGE}}$ value of 0.74 ($\pm$0.12), equivalent to using more than 6 random samples. Additional results (reported in the Appendix) confirm that a similar advantage is also found when measuring Spearman's rank correlation ($\rho$) instead of Kendall's $\tau$.

\paragraph{Ability to rank exceptionally high-quality explanations}

\begin{figure}[htbp]
    \centering
    \includegraphics[width=\columnwidth]{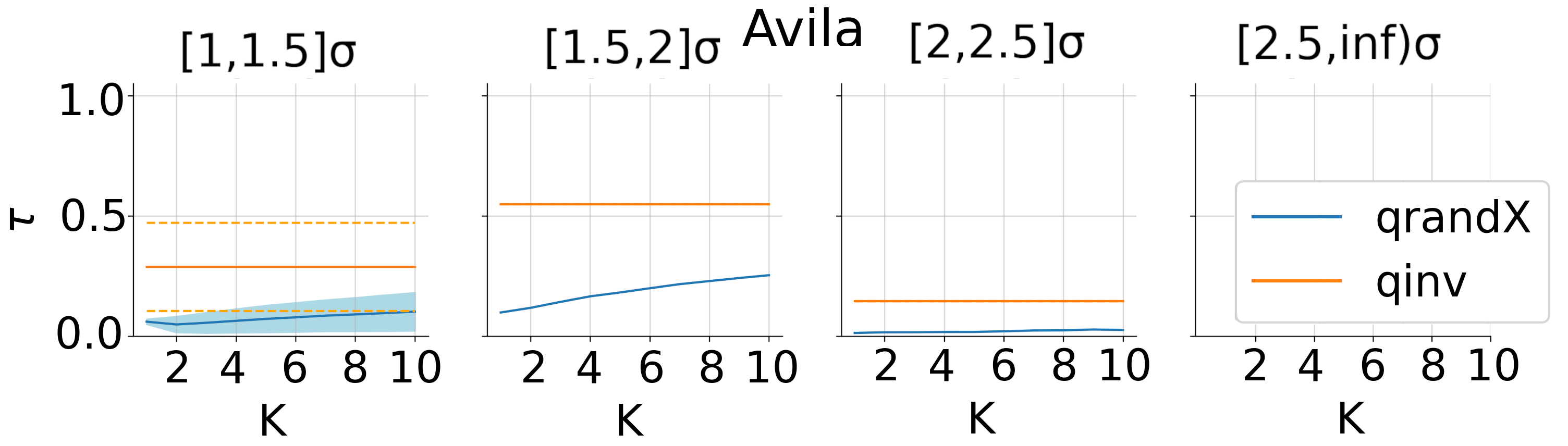}
    \includegraphics[width=\columnwidth]{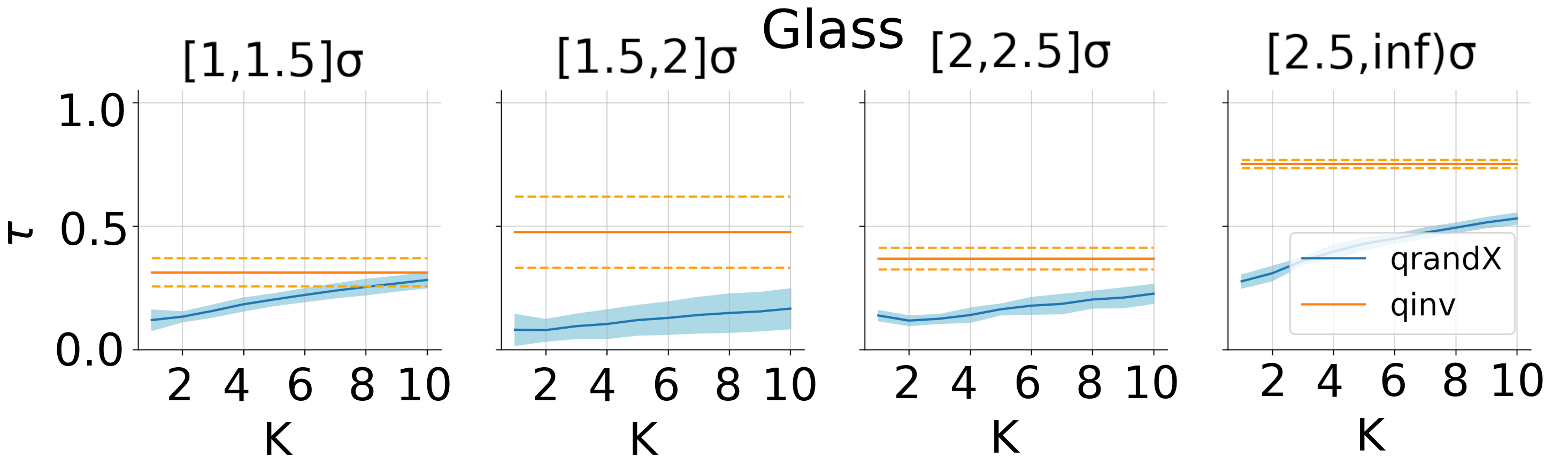}
    \caption{Kendall's $\tau$ for explanations of a given level of exceptionality. The blue line represents the average correlation $\tau_{q,\texttt{QRAND}_{\text{K}}}$ for each $K$ value across 10 different inputs, with the shaded area indicating the average$\pm\sigma$. The orange line shows the average correlation $\tau_{q,\texttt{QGE}}$,  with dashed lines marking the average $\pm\sigma$.}
    \label{fig:exhaustive_tau_exceptional}
\end{figure}

We analyzed Kendall's $\tau$ between $q$ and $qt$ for different subsets of explanations, stratified by their quality levels. The results, depicted in Fig.~\ref{fig:exhaustive_tau_exceptional}, illustrate that the capability of $qt$ to accurately rank explanations improves with the increasing quality of the explanations. Notably, this improvement is significantly more pronounced for $qt=\texttt{QGE}$ compared to $qt=\texttt{QRAND}_{\text{K}}$, indicating a superior performance in distinguishing high-quality explanations, which are usually the focus of practitioners.

These experiments show that \texttt{QGE} meets requirement \textbf{R2} to a higher extent than the existing alternative, $\texttt{QRAND}_{\text{K}}$ except for large values of $K$. Moreover, the computational cost of $\texttt{QRAND}_{\text{K}}$ is proportional to $K$, so it needs to far exceed the computational cost of \texttt{QGE}, which is on par with the computational cost of $\texttt{QRAND}_1$ (two calculations of $\Psi$). This demonstrates that \texttt{QGE} also complies with \textbf{R3} to a much higher extent than $\texttt{QRAND}_{\text{K}}$ to obtain comparable performance.

\subsubsection{b. Performance on larger datasets}\label{sec:large_dataset_exploration}
To confirm the applicability of our findings across different types of data and larger datasets, we expanded our experiments to include both image datasets (\texttt{MNIST}, \texttt{CIFAR}, \texttt{ImageNet}) and a textual dataset (\texttt{20newsgroups}).


For \texttt{20newsgroups}, we trained an \texttt{MLP} model that achieved an accuracy of 0.78. For \texttt{MNIST}, we utilized a small convolutional network with 0.93 accuracy, and for \texttt{CIFAR}, a \texttt{ResNet50} network obtained an accuracy of 0.77. With \texttt{ImageNet}, we tested five different pre-trained models from the TorchVision library: three convolutional architectures (\texttt{ResNet18}, \texttt{ResNet50}, and \texttt{VGG16}) and two attention-based architectures (\texttt{ViT\_b\_32} and \texttt{MaxViT\_t}).

Due to the impracticality of processing all possible explanations for these larger datasets, we sampled 10,000 random explanations for each tested input\footnote{Although random explanations frequently have lower quality than explanations obtained with existing explanation methods, the latter are costly to obtain, and too few to yield statistically significant results. Moreover, Section~\ref{sec:exhaustive_exploration} shows that the effect observed for average-quality explanations is maintained or enhanced for high-quality explanations, indicating that exploring a sizable set of random explanations is more informative for this experiment than exploring a handful of very good explanations.}. For each dataset-model pair, 25 different inputs were tested, and we report the average increase in Kendall's $\tau$ when using \texttt{QGE} as opposed to $\texttt{QRAND}_1$ ($\Delta \tau=\tau_{q,\texttt{QGE}}-\tau_{q,\texttt{QRAND}_1}$) and the average 
 increase in Spearman's $\rho$ ($\Delta\rho=\rho_{q,\texttt{QGE}}-\rho_{q,\texttt{QRAND}_1}$).


The results\footnote{Importantly, a well-known issue when explaining predictions involving images is that pixel-level perturbations interact suboptimally with convolutional networks. A prevalent solution involves using explanations at the superpixel level~\cite{blücher2024decoupling}). Therefore, we used superpixel level explanations for \texttt{CIFAR} and \texttt{ImageNet} (4x4 and 32x32 superpixels, respectively). For completeness, the results for explanations at the pixel level are reported in the Appendix.}, detailed in Table \ref{tab:effects_per_model_and_dataset_chunky}, consistently show a substantial positive impact on both Kendall and Spearman correlation when using \texttt{QGE} instead of $\texttt{QRAND}_1$ across a variety of datasets and model architectures.

\begin{table}[!t]
    \vskip 0.15in
    \begin{center}
    \begin{small}
    \begin{sc}
    \begin{tabular}{l|l|c|c}
        Dataset & Model & $\Delta \tau$ & $\Delta\rho$\\
        \toprule
        \texttt{20newsgroups} & \texttt{mlp} & 0.090$\pm$0.05 & 0.093$\pm$0.06\\
        \texttt{mnist} & \texttt{mlp} & 0.214$\pm$0.04 & 0.200$\pm$0.03\\
        \texttt{cifar} & \texttt{resnet50} & 0.054$\pm$0.04 & 0.059$\pm$0.05\\
        \texttt{imagenet} & \texttt{resnet18} & 0.110$\pm$0.02 & 0.115$\pm$0.02\\
        \texttt{imagenet} & \texttt{resnet50} & 0.150$\pm$0.03 & 0.149$\pm$0.03\\
        \texttt{imagenet} & \texttt{vgg16} & 0.108$\pm$0.08 & 0.106$\pm$0.07\\
        \texttt{imagenet} & \texttt{vit\_b\_32} & 0.139$\pm$0.03 & 0.141$\pm$0.02\\
        \texttt{imagenet} & \texttt{maxvit\_t} & 0.132$\pm$0.08 & 0.127$\pm$0.07\\
        \bottomrule
    \end{tabular}
    \end{sc}
    \end{small}
    \end{center}
    \caption{Magnitude of the increase in Kendall and Spearman correlation when using \texttt{QGE} instead of $\texttt{QRAND}_1$ on large datasets.}\label{tab:effects_per_model_and_dataset_chunky}
\end{table}

\subsubsection{c. Advantage of using \texttt{QGE} vs. $\texttt{QRAND}_1$ across datasets and models}
\label{sec:experiments_datasets_models}
As discussed in Section \ref{sec:suitability_qge}, using the \texttt{Pixel-Flipping} average activation as a quality measure necessitates a perturbation function that transforms inputs, potentially pushing the model to operate on data points outside of its training distribution~\cite{hase2021}. To mitigate any effects from this interaction, we repeated the experiments from Section \ref{sec:exhaustive_exploration} using models that were exposed to masked inputs during training, as outlined in~\cite{hase2021}. Furthermore, we experimented with masking using the average value for each input attribute, rather than zeros. We also tested models with reduced accuracies to diversify the conditions. Table \ref{tab:effects_per_model} summarizes the average results from these experiments for five different inputs.

\begin{table}[!t]
    \vskip 0.15in
    \begin{center}
    \begin{small}
    \begin{sc}
    \begin{tabular}{l|c|c|c|c}
        \multicolumn{5}{c}{{Avila}} \\
        \hline
        Model & Acc. & $\sigma(q)$ &$\Delta \tau$ & $\Delta\rho$\\
        \toprule
        \texttt{MLP} & 0.99 & 0.191 & 0.198$\pm$0.12 & 0.173$\pm$0.10\\
        \texttt{ood-mean} & 0.80 & 0.080 & 0.288$\pm$0.05 & 0.245$\pm$0.03\\
        \texttt{ood-zeros} & 0.80 & 0.085 & 0.247$\pm$0.09 & 0.213$\pm$0.06\\
        \texttt{undertr.} & 0.75 & 0.105 & 0.315$\pm$0.06 & 0.257$\pm$0.03\\
        \texttt{untrained} & 0.05 & 0.001 & 0.434$\pm$0.01 & 0.298$\pm$0.00\\
        \hline
        \multicolumn{5}{c}{{Glass}}\\
        \hline
        Model & Acc. & $\sigma(q)$ &$\Delta \tau$ & $\Delta\rho$\\
        \toprule
        \texttt{MLP} & 0.77 & 0.198 & 0.241$\pm$0.11 & 0.204$\pm$0.09\\
        \texttt{ood-mean} & 0.63 & 0.070 & 0.314$\pm$0.01 & 0.262$\pm$0.01\\
        \texttt{ood-zeros} & 0.63 & 0.052 & 0.267$\pm$0.06 & 0.230$\pm$0.04\\
        \texttt{undertr}. & 0.60 & 0.168 & 0.191$\pm$0.02 & 0.184$\pm$0.02\\
        \texttt{untrained} & 0.23 & 0.007 & 0.173$\pm$0.10 & 0.163$\pm$0.08\\
        \hline
    \end{tabular}
    \end{sc}
    \end{small}
    \end{center}
    \caption{Magnitude of the increase in Kendall and Spearman correlation when using \texttt{QGE} instead of $\texttt{QRAND}_1$ for the \texttt{Avila} and \texttt{Glass} datasets and different models all using an \texttt{MLP} architecture: \texttt{MLP} refers to the fully trained model exposed to no masked input; \texttt{ood-mean} and \texttt{ood-zeros} were exposed during training to inputs masked with zeros and the average value of each attribute, respectively; \texttt{undertr.} was trained only until achieving 70\% the accuracy of the fully trained model; and \texttt{untrained} was not exposed to any data. $\sigma(q)$ indicates the standard deviation of the distribution of $q$ across all possible explanations.}\label{tab:effects_per_model}
\end{table}

These results show that using \texttt{QGE} consistently yields better outcomes than $\texttt{QRAND}_1$, irrespective of the dataset or model type. However, the nature of the model significantly influences the extent of the advantage offered by \texttt{QGE}. A deeper analysis of this effect is included in Appendix~\ref{sec:exploring-std}.

\subsubsection{d. Suitability for other quality metrics}\label{sec:suitability_for_metrics}
All experiments reported above use \texttt{Pixel-Flipping} \cite{bach2015pixel} as a quality metric. This metric is a popular choice, which is why we have explored it extensively. However, to determine whether the observed effects are consistent across different quality measures, we tested our method using a variety of measures spanning different quality dimensions. For all measures, we used the implementations in the Quantus~\cite{hedstrom2022quantus} library.

\paragraph{Faithfulness metrics}
In addition to \texttt{Pixel-Flipping}, we tested \texttt{Faithfulness Correlation} \cite{bhatt2020}, \texttt{Faithfulness Estimate} \cite{AlvarezCorr18}, and \texttt{Monotonicity Correlation} \cite{arya2019explanation} for 1,000 random explanations. The results, summarized in Table \ref{tab:effects_faithfulness}, show that \texttt{QGE} performs superiorly to $\texttt{QRAND}_1$ for all measures, obtaining substantial increases for three of the four measures. However, it offers little advantage for \texttt{Faithfulness Correlation}. This metric is known to be unstable, often yielding highly variable results for the same input across different executions \cite{tomsett2020, hedstrom2023metaquantus, hedstrom2022quantus}. This variability undermines the informative advantage of $\Psi(\mathbf{e^{inv}}, \mathbf{x}, f, \hat{y})$ (Eq. \ref{eq:qge}) over $\Psi(\mathbf{e^{r}}, \mathbf{x}, f, \hat{y})$ using a random explanation $\mathbf{e^{r}}$, explaining the lack of advantage observed.

\begin{table}[!t]
    \vskip 0.15in
    \begin{center}
    \begin{small}
    \begin{sc}
    \begin{tabular}{l|c|c}
        \multicolumn{3}{c}{\texttt{Pixel-Flipping}}\\
        \hline
        Model & $\Delta \tau$ & $\Delta \rho$\\
        \toprule
        \texttt{ResNet18} & 0.142$\pm$0.07 & 0.138$\pm$0.06\\
        \texttt{VGG16} & 0.102$\pm$0.04 & 0.106$\pm$0.03\\
        \hline
        \multicolumn{3}{c}{\texttt{FaithfulnessCorrelation}}\\
        \hline
        Model & $\Delta PA$ & $\Delta \rho$\\
        \toprule
        \texttt{ResNet18} & 0.007$\pm$0.01 & 0.007$\pm$0.02\\
        \texttt{VGG16} & 0.009$\pm$0.01 & 0.012$\pm$0.01\\
        \hline
        \multicolumn{3}{c}{\texttt{FaithfulnessEstimate}}\\
        \hline
        Model & $\Delta PA$ & $\Delta \rho$\\
        \toprule
        \texttt{ResNet18} & 0.368$\pm$0.02 & 0.279$\pm$0.02\\
        \texttt{VGG16} & 0.397$\pm$0.02 & 0.290$\pm$0.02\\
        \hline
        \multicolumn{3}{c}{\texttt{MonotonicityCorrelation}}\\
        \hline
        Model & $\Delta PA$ & $\Delta \rho$\\
        \toprule
        \texttt{ResNet18} & 0.353$\pm$0.05 & 0.277$\pm$0.02\\
        \texttt{VGG1} & 0.374$\pm$0.03 & 0.288$\pm$0.01\\
        \hline
    \end{tabular}
    \end{sc}
    \end{small}
    \end{center}
    \caption{Magnitude of the increase in Kendall and Spearman correlation when using \texttt{QGE} instead of $\texttt{QRAND}_1$ for faithfulness metrics on predictions of a \texttt{ResNet18} and \texttt{VGG16} models on the \texttt{Imagenet} dataset.}
    \label{tab:effects_faithfulness}
\end{table}

\paragraph{Localization metrics}
We evaluated the effectiveness of \texttt{QGE} using various localization measures, including \texttt{AttributionLocalisation} \cite{kohlbrenner2020towards}, \texttt{TopKIntersection} \cite{theiner2021}, \texttt{RelevanceRankAccuracy}, \texttt{RelevanceMassAccuracy} \cite{arras2021ground}, and \texttt{AUC}~\cite{Fawcett}. For these tests, we utilized the \texttt{CMNIST} dataset \cite{bykov2021noisegrad}, training a \texttt{ResNet18} model to perform the evaluations. We then assessed the localization of 10,000 random attributions using these measures, applying both the $\texttt{QRAND}_1$ and \texttt{QGE} transformations. The results are summarized in Table \ref{tab:effects_localization}, which reports increases in Kendall and Spearman correlation when using \texttt{QGE} instead of $\texttt{QRAND}_1$. \texttt{QGE} consistently outperforms $\texttt{QRAND}_1$ across all tested localization metrics, enhancing both the Kendall and the Spearman correlation of the transformed metrics with the original ones. The magnitude of the advantage that \texttt{QGE} provides over $\texttt{QRAND}_1$ varies depending on the nature of the quality metric used, with the most significant improvements observed using \texttt{AttributionLocalisation}, \texttt{RelevanceMassAccuracy} and \texttt{AUC}.

\begin{table}[!t]
    \vskip 0.15in
    \begin{center}
    \begin{small}
    \begin{sc}
    \begin{tabular}{l|c|c}
        \multicolumn{3}{c}{\texttt{CMNIST} - \texttt{ResNet18}}\\
        \toprule
        Measure &$\Delta \tau$ & $\Delta\rho$\\
        \hline
        \texttt{Attr.Loc.} & 0.500$\pm$0.00 & 0.309$\pm$0.00\\
        \texttt{TopKInt.} & 0.015$\pm$0.01 & 0.022$\pm$0.01\\
        \texttt{Rel.RankAcc.} & 0.089$\pm$0.02 & 0.093$\pm$0.02\\
        \texttt{Rel.MassAcc.} & 0.501$\pm$0.01 & 0.310$\pm$0.01\\
        \texttt{AUC} & 0.496$\pm$0.00 & 0.304$\pm$0.00\\
        \hline
    \end{tabular}
    \end{sc}
    \end{small}
    \end{center}
    \caption{Increase in Kendall and Spearman correlation when using \texttt{QGE} instead of $\texttt{QRAND}_1$ for different localization metrics 
    on predictions of a \texttt{ResNet18} model on \texttt{CMNIST} data.}
    \label{tab:effects_localization}
\end{table}

\paragraph{Robustness and randomization metrics}
We considered incorporating robustness and randomization metrics into our evaluations. These metrics assess the explanation method itself rather than the explanations obtained. While they can be quantified using the \texttt{QGE} transformation, they are not amenable to comparison using $\texttt{QRAND}_1$, as the latter does not rely on the explanation method. Therefore, no direct comparison between \texttt{QGE} and $\texttt{QRAND}_1$ was feasible.

\subsection{Effect on Existing Quality Metrics}\label{sec:effect_on_metrics}

We investigate the statistical reliability of the \texttt{QGE} transformation compared to the original quality metric. 
We follow the meta-evaluation methodology outlined in \cite{hedstrom2023metaquantus} where metric reliability is assessed in two steps: performing a minor- (noise resilience, $NR$) or disruptive- (reactivity to adversaries, $AR$) perturbation and then, measuring how the metric scores and method rankings changed, post-perturbation. 
For each perturbation scenario, intra-consistency ($\text{IAC}$; the similarity of score distributions under perturbation), and inter-consistency ($\text{IEC}$; the ranking similarity among different explanation methods) are computed, resulting in a meta-consistency vector $\mathbf{m} \in \mathbb{R}^{4}$ and a summarising score $\mathbf{MC} \in [0, 1]$:
\begin{equation}
    \mathbf{MC} = \left(\frac{1}{|\mathbf{m^{*}}|}\right){\mathbf{m^{*}}}^T\mathbf{m} \quad \text{where} \quad \mathbf{m} = \begin{bmatrix} 
    {IAC}_{NR}\\
    {IAC}_{AR}\\
    {IEC}_{NR}\\ 
    {IEC}_{AR}
    \end{bmatrix},
    \label{eq:meta-eval-vector}
\end{equation}
where $\mathbf{m}^{*} = \mathbb{1}^4$
represents an optimally performing quality estimator. 
A higher $\mathbf{MC}$ score, approaching 1, signifies greater reliability on the tested criteria. Perturbations are applied to either the model parameters or input.


We used the \texttt{fMNIST} \cite{fashionmnist2015} and \texttt{ImageNet} \cite{ILSVRC15} datasets. For the first toy dataset we use the LeNet architecture \cite{lecun2010mnist} and for \texttt{ImageNet} we use a pre-trained \texttt{ResNet-18} \cite{he2015deep} from \cite{Pytorch2019}. The results shown in Table~\ref{tab:effects_existing_measures} demonstrate that the \texttt{QGE} method yields reliability improvements across tested metrics. This performance enhancement is most notable for \texttt{Pixel-Flipping}, where \texttt{QGE} significantly enhances the inter-consistency (IEC) under adversarial reactivity (\textit{AR}), indicating a marked improvement in the metric's ability to differentiate between meaningful and random inputs and models, as detailed in Appendix~\ref{sec:effect-on-existing-appendix}. Also, \texttt{QGE}'s effect on localization is competitive, though not statistically significant.

\begin{table}[!t]
    \begin{center}
    \begin{small}
    \begin{sc}
    \begin{tabular}{l|c|c|c}
        \multicolumn{4}{c}{\texttt{fMNIST} - \texttt{LeNet}}\\
        \toprule
        Measure & PF & RRA & RMA\\
        \hline
        $\texttt{QRAND}_1$ & 0.579 & 0.599 & 0.585\\
        \texttt{QGE} & 0.801 & 0.598 & 0.587\\
        \hline
        \multicolumn{4}{c}{\texttt{ImageNet} - \texttt{ResNet18}}\\
        \toprule
        Measure & PF & RRA & RMA\\
        \hline
        $\texttt{QRAND}_1$ & 0.634 & 0.594 & 0.596\\
        \texttt{QGE} & 0.849 & 0.568 & 0.619\\
        \hline
    \end{tabular}
    \end{sc}
    \end{small}
    \end{center}
    \caption{MC score when using \texttt{QGE} and $\texttt{QRAND}_1$ for \texttt{Pixel-Flipping} (PF) (Faithfulness), \texttt{Relevance Rank Accuracy} (RRA) and \texttt{Relevance Mass Accuracy} (RMA) (Localization) on predictions of a \texttt{ResNet18} model on the \texttt{fMNIST} dataset.}
    \label{tab:effects_existing_measures}
\end{table}
\section{Conclusions and Future Work}
\label{sec:Conclusions}
In this work, we introduced the Quality Gap Estimator (\texttt{QGE}), designed to compare the quality of an explanation against alternative explanations, aiding practitioners in determining the need to seek better alternatives. \texttt{QGE} is computationally efficient and can be used with most quality metrics commonly used in XAI, improving their informativeness.

By conceptualizing the challenge of achieving a relative quality measurement as a sampling issue, we demonstrated that \texttt{QGE} is more sample-efficient than the conventional method of comparing with a single random explanation. Extensive testing across various datasets, models, and quality metrics has consistently shown that employing \texttt{QGE} is advantageous over the traditional approach.

Additionally, the transformation implemented by \texttt{QGE} results in quality metrics with enhanced statistical significance, suggesting its utility even in scenarios where relative comparisons are not the primary objective.

For future work, we aim to enhance \texttt{QGE}'s performance with metrics that are inherently unstable, where it currently does not offer a significant improvement over the comparison with a single random sample. Further, we are interested in exploring the potential of employing a similar strategy to also improve the explanations, extending the utility of \texttt{QGE} beyond mere quality measurement.

\appendix
\section{Additional Results}
This Appendix lists results that complement those mentioned in Section~\ref{sec:experimental-results}.

\subsection{Distribution of QGE}\label{sec:distribution-qge}
\begin{figure}[htbp]
    \centering
    \includegraphics[width=0.5\linewidth]{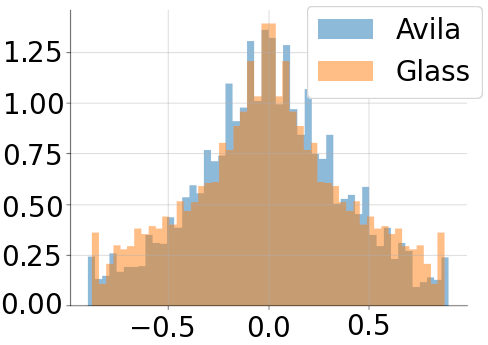}
    \caption{Density histogram of \texttt{QGE} for every possible explanation of 5 different input samples for datasets \texttt{Avila} and \texttt{Glass}.}
    \label{fig:qge_distribution}
\end{figure}

The exhaustive exploration of all possible explanations for the 5 input samples used for each of the \texttt{Avila} and \texttt{Glass} datasets confirms that the distribution of \texttt{QGE} is centered around zero, as desired for \textbf{R1} and shown in Fig.~\ref{fig:qge_distribution}.

\subsection{Spearman Correlation Results}\label{sec:exhaustive-spearman}
\begin{figure}[htbp]
    \centering
    \includegraphics[width=0.45\linewidth]{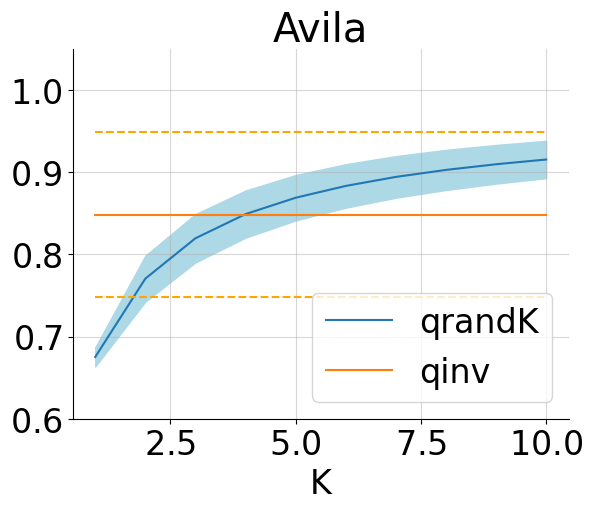}
    \includegraphics[width=0.45\linewidth]{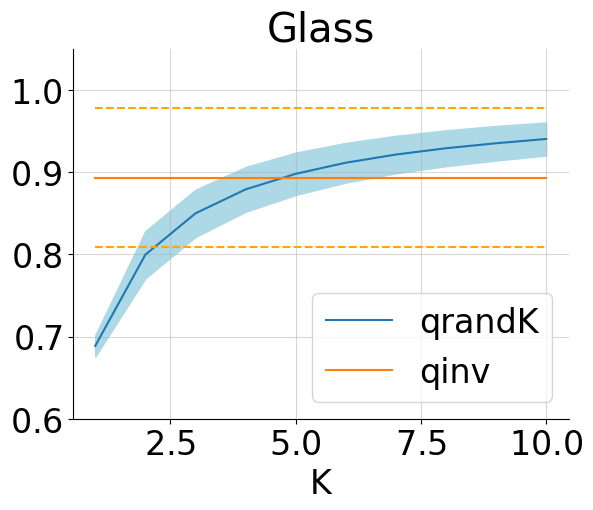}
    \caption{Spearman correlation of $qt$ with the original $q$ for the \texttt{Avila} and \texttt{Glass} datasets. The blue line indicates the average correlation $\rho_{q,\texttt{QRAND}_{\text{K}}}$ for each value of $K$ over 5 different inputs. The shaded area shows the average $\pm\sigma$. The orange line records the average correlation $\rho_{q,\texttt{QGE}}$ with dashed lines representing the average $\pm\sigma$.}
    \label{fig:exhaustive_spearman}
\end{figure}

\normalsize
To complete the analysis in Section~\ref{sec:large_dataset_exploration}.a we also measured the Spearman correlation ($\rho_{q,qt}$) across the same 5 input samples. Fig.~\ref{fig:exhaustive_spearman} shows that \texttt{QGE} outperforms $\texttt{QRAND}_{\text{K}}$ with up to $k=4$ samples for \texttt{Avila}, and $k=5$ for \texttt{Glass}.

\subsection{Pixel-Level Explanations}\label{sec:pixel-level-explanations}
Section~\ref{sec:large_dataset_exploration}.b reports experiments performed on superpixel-level explanations. For those explanations, instead of perturbing input variables separately, the perturbations are applied to blocks of contiguous pixels, denominated superpixels. The superpixel size used for \texttt{CIFAR} was 4x4, while for \texttt{ImageNet} we used 32x32. For completeness, Table~\ref{tab:effects_per_model_and_dataset} reports the results for perturbations applied to individual pixels, which is analogous to the perturbation mode used in all other datasets. These results show that despite having less of an impact than for super-pixel-level explanations, using \texttt{QGE} is advantageous over $\texttt{QRAND}_1$. The enhancement of the effect for superpixel-level explanations (which are higher-quality explanations), confirms the result listed in Section~\ref{sec:exhaustive_exploration} that indicates that the advantage of \texttt{QGE} is larger the higher the quality of the explanations.

\begin{table}[!t]
    \vskip 0.15in
    \begin{center}
    \begin{small}
    \begin{sc}
    \begin{tabular}{l|l|c|c}
        Dataset & Model & $\Delta \tau$ & $\Delta\rho$\\
        \toprule
        \texttt{cifar} & \texttt{resnet50} & 0.054$\pm$0.03 & 0.058$\pm$0.03\\
        \texttt{imagenet} & \texttt{resnet18} & 0.018$\pm$0.01 & 0.020$\pm$0.01\\
        \texttt{imagenet} & \texttt{resnet50} & 0.019$\pm$0.01 & 0.021$\pm$0.01\\
        \texttt{imagenet} & \texttt{vgg16} & 0.016$\pm$0.01 & 0.018$\pm$0.01\\
        \texttt{imagenet} & \texttt{vit\_b\_32} & 0.040$\pm$0.02 & 0.043$\pm$0.02\\
        \texttt{imagenet} & \texttt{maxvit\_t} & 0.017$\pm$0.02 & 0.018$\pm$0.02\\
        \bottomrule
    \end{tabular}
    \caption{Magnitude of the increase in Kendall and Spearman correlation when using \texttt{QGE} instead of $\texttt{QRAND}_1$ for explanations at the pixel level.}
    \label{tab:effects_per_model_and_dataset}
    \end{sc}
    \end{small}
    \end{center}
\end{table}

\subsection{Variation of the Effect With the Distribution's Standard Deviation}\label{sec:exploring-std}

\begin{figure}[htbp]
    \centering
    \includegraphics[width=0.33\columnwidth]{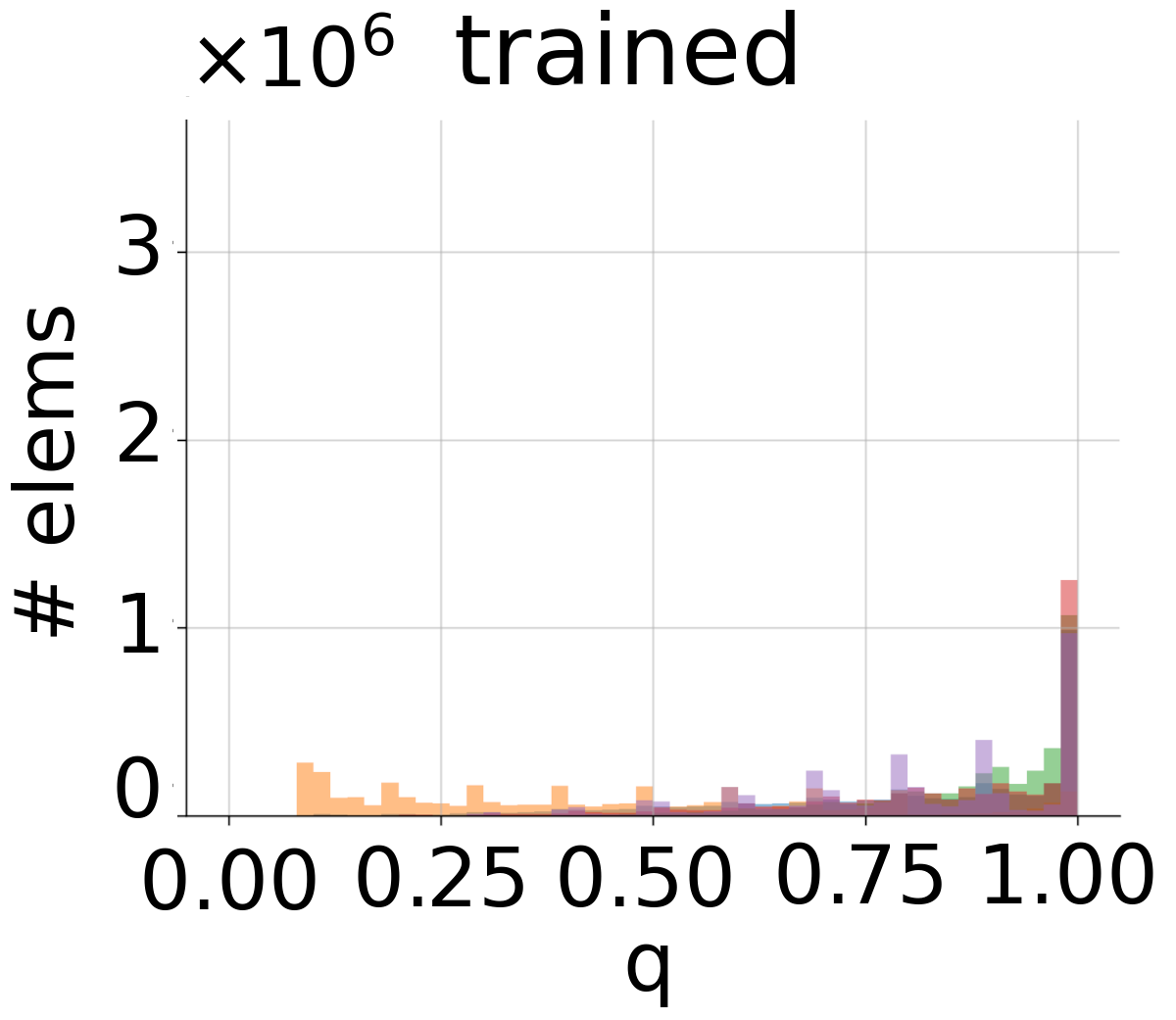}
    \includegraphics[width=0.30\columnwidth]{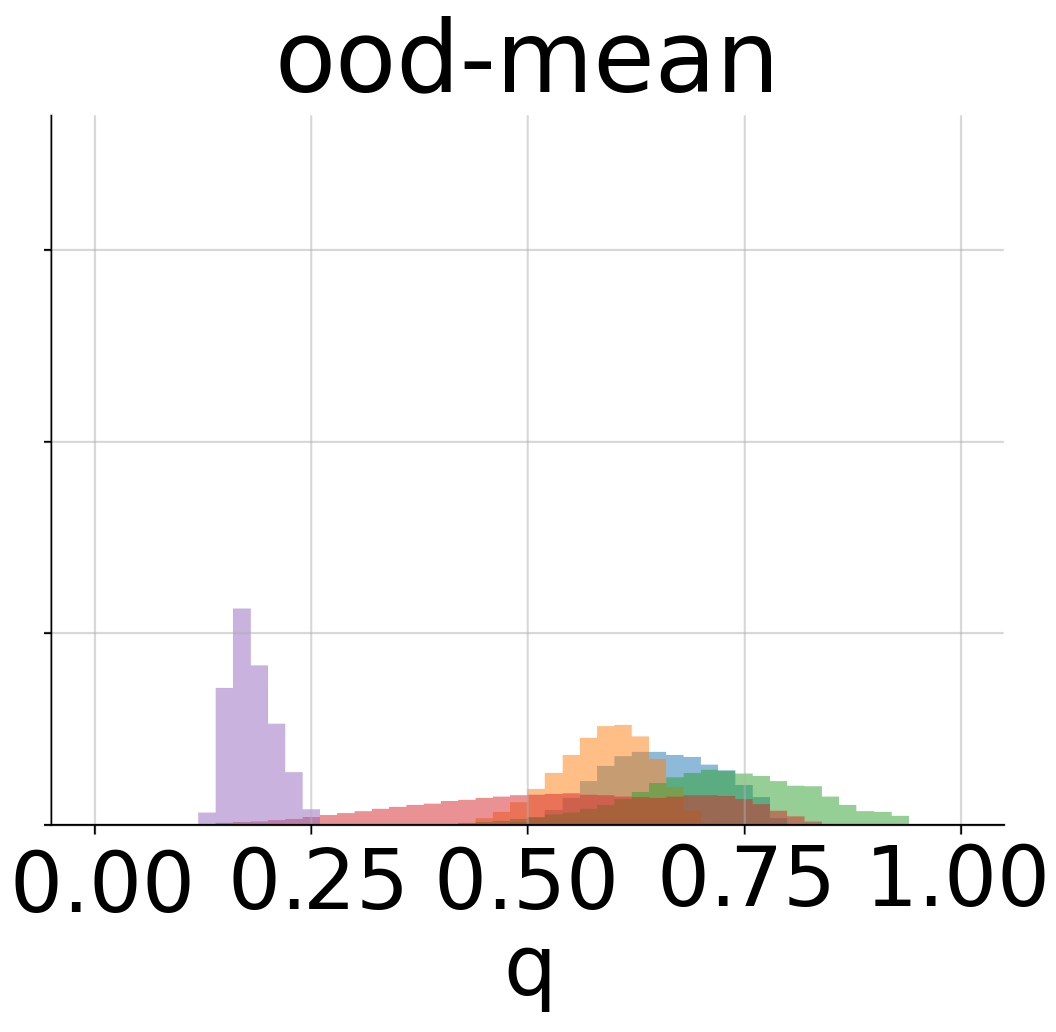}
    \includegraphics[width=0.30\columnwidth]{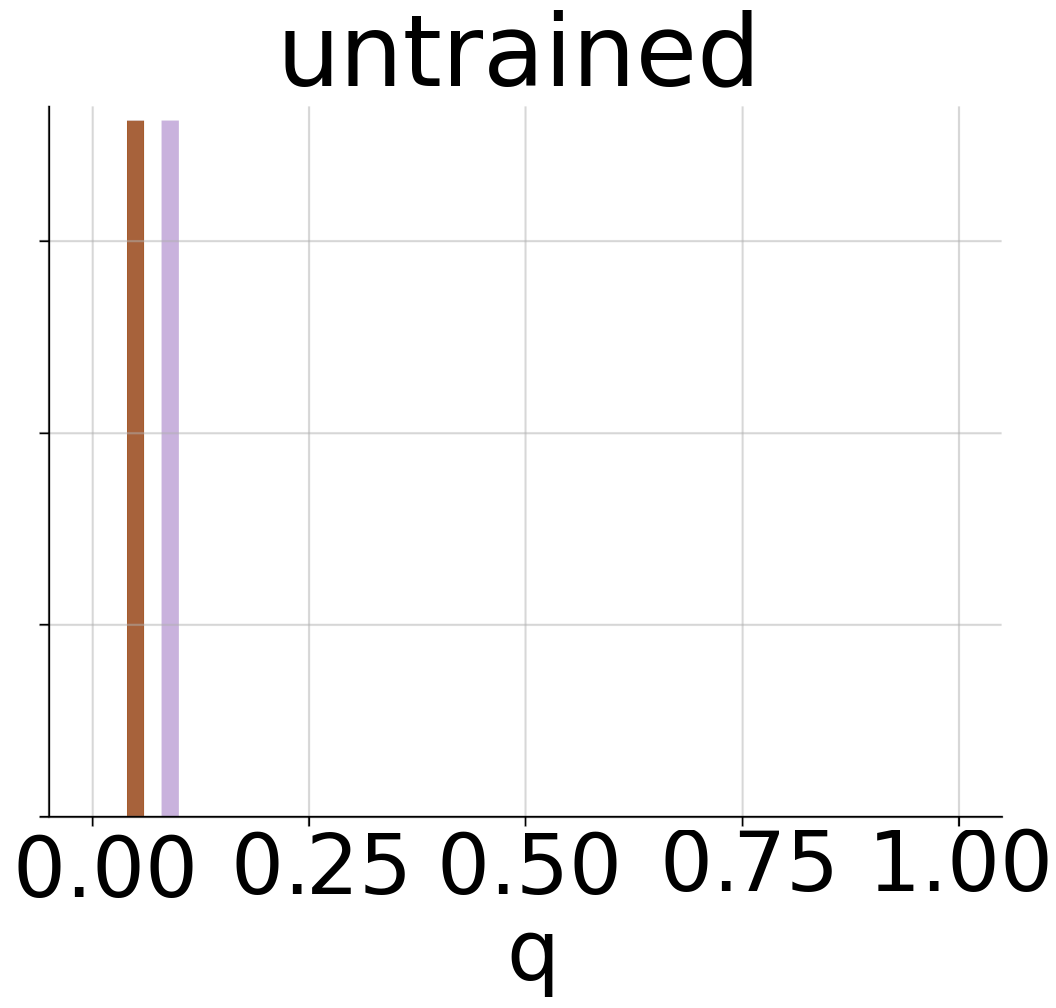}
    \caption{Histograms for the distributions of $q$ for 5 different inputs using the \textit{trained}, \textit{ood-mean} and \textit{untrained} models.}
    \label{fig:exhaustive_qmeans_hists}
\end{figure}

A deeper analysis of the results in Section~\ref{sec:experiments_datasets_models}.c shows considerable variation in the distribution of $q$ depending on the model used, as illustrated in Figure \ref{fig:exhaustive_qmeans_hists}. Table \ref{tab:effects_per_model} also presents the average standard deviation of the $q$ distribution for each model. The fully-trained \textit{mlp} model exhibits a wide range of $q$ values for each input (the average $\sigma$ is 0.19), indicating a substantial difference in quality between the best and worst explanations. In contrast, the \textit{undertrained} model displays significantly less variability in $q$ across both datasets. The most pronounced case is the \textit{untrained} model, which shows highly concentrated $q$ distributions, i.e., minimal numerical differences between the $q$ values of the best and worst explanations. Despite these variations, the impact of utilizing \texttt{QGE} over $\texttt{QRAND}_1$ is consistently positive, confirming its robustness and effectiveness across various conditions.

\subsection{Effect on Existing Quality Measures}\label{sec:effect-on-existing-appendix}

\begin{figure}[h!]
    \begin{subfigure}[b]{\columnwidth}
        \centering
        \includegraphics[width=\columnwidth]{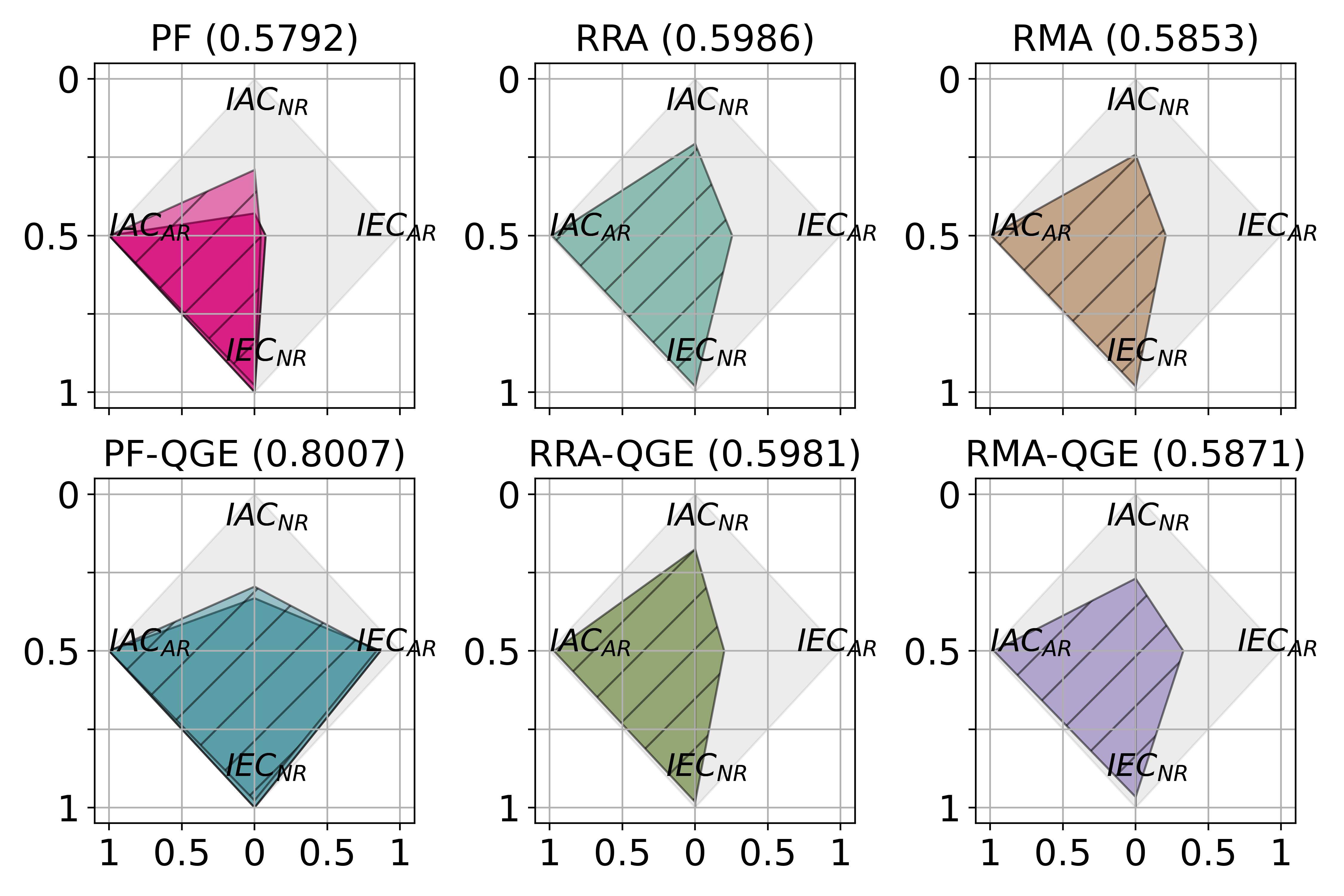}
        \caption{\texttt{fMNIST - LeNet}}
        \label{fig:fmnist}
    \end{subfigure}
    
    \vspace{1em} 
    
    \begin{subfigure}[b]{\columnwidth}
        \centering
        \includegraphics[width=\columnwidth]{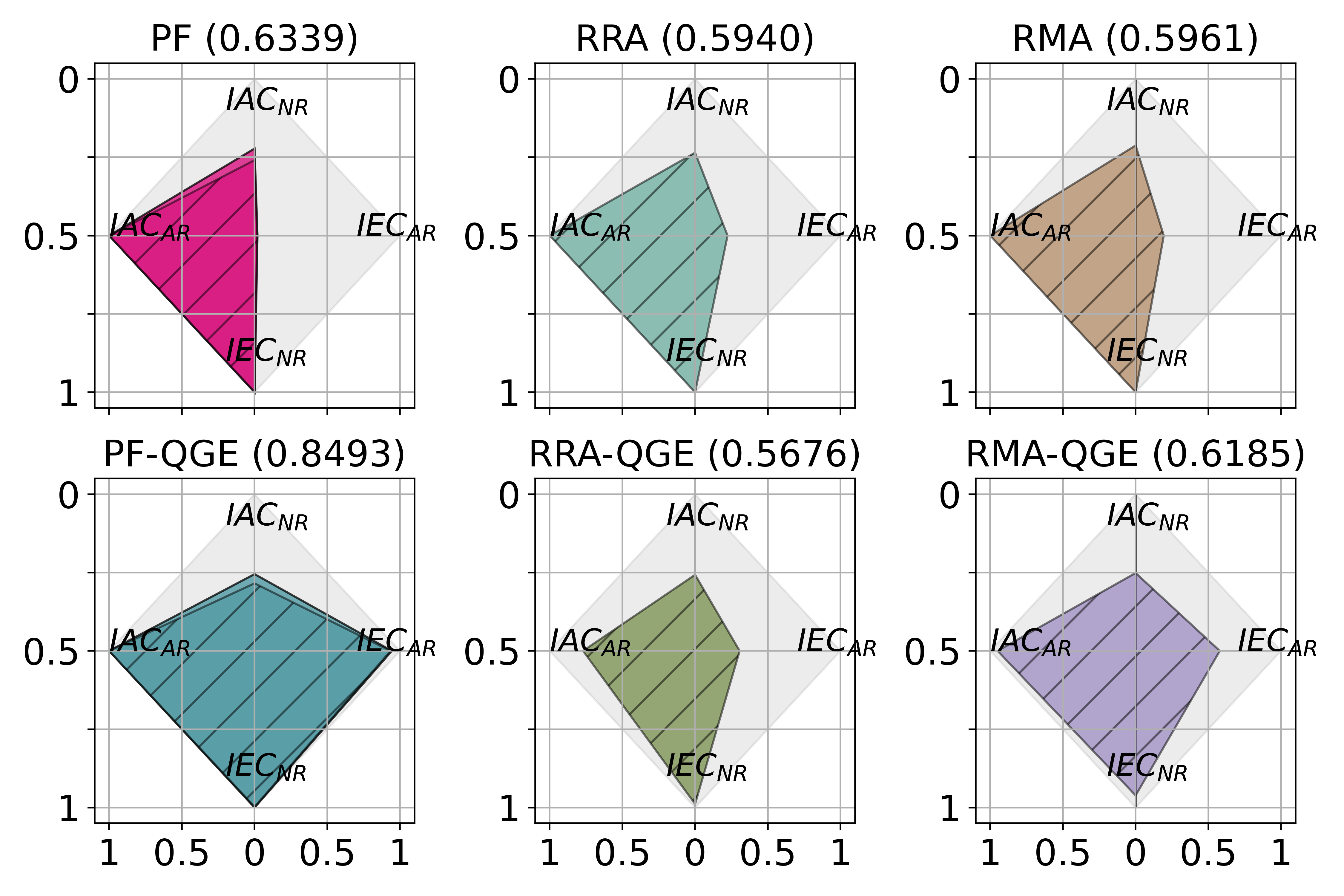}
        \caption{\texttt{ImageNet - ResNet18}}
        \label{fig:imagenet}
    \end{subfigure}
    \caption{
 A graphical representation of the benchmarking results aggregated over 3 iterations with $K=5$. We use $\{\textit{Saliency, Integrated Gradients, Input X Gradient}\}$ as explanation methods.
 Each column corresponds to a quality estimator, from left to right: \texttt{Pixel-Flipping} (PF) (Faithfulness), \texttt{Relevance Rank Accuracy} (RRA) and \texttt{Relevance Mass Accuracy} (RMA)
 (Localization). The bottom row shows results with \texttt{QGE}. Solid shapes correspond to input perturbations, and striped shapes to model perturbations. The grey area indicates the area of an optimally performing estimator, i.e., $\mathbf{m}^{*} = \mathbb{1}^4$. The MC score (indicated in brackets) is averaged over Model- and Input perturbation tests. Higher values and larger colored areas indicate higher performance.} 
\label{fig:meta_eval_area_graph_fmnist}
\end{figure}

In Figure \ref{fig:meta_eval_area_graph_fmnist}, we show the different area graphs which each contain the results from the meta-evaluation analysis (set as coordinates on a 2D plane) for the \texttt{fMNIST} \cite{fashionmnist2015} and \texttt{ImageNet} \cite{ILSVRC15} datasets, respectively. The titles hold the summarising $\mathbf{MC}$ score and each edge contains the meta-evaluation vector scores. By inspecting the colored areas of the respective estimators in terms of their size and shape, we can deduce the overall performance of both failure modes. Larger colored areas imply better performance on the different scoring criteria and the grey area indicates the area of an optimally performing quality estimator. The Quantus \cite{hedstrom2022quantus} and MetaQuantus \cite{hedstrom2023metaquantus} libraries were used for the experiments.

\section{Code and Reproducibility}
An implementation of \texttt{QGE} for a wide variety of quality measures is available in the Quantus toolkit: \url{https://github.com/understandable-machine-intelligence-lab/Quantus}

The code used for all experiments is available at \url{https://github.com/annahedstroem/eval-project}

\ifdefined\doubleblind
\else
\section*{Acknowledgments}
Carlos Eiras-Franco wishes to thank CITIC, which financed the research stay that originated this work. As a Research Center accredited by the Galician University System, CITIC is funded by ``Consellería de Cultura, Educación e Universidades from Xunta de Galicia", supported in an 80\% through ERDF Funds, ERDF Operational Programme Galicia 2014-2020, and the remaining 20\% by ``Secretaría Xeral de Universidades" (Grant ED431G 2019/01). This publication is part of project PID2021-128045OA-I00, financed by MCIN/AEI/10.13039/501100011033/FEDER, UE.
Carlos Eiras-Franco also thanks the support received by the Xunta de Galicia (Grant ED431C 2018/34) with the European Union ERDF funds, and the support received by the National Plan for Scientific and Technical Research and Innovation of the Spanish Government (Grant PID2019-109238GBC22) and the European Union ERDF funds.
This work was funded by the German Ministry for Education and Research through project Explaining 4.0 (ref. 01IS200551).
\fi

\small
\bibliography{references}

\end{document}